%% file: main.tex
\pdfoutput=1
\documentclass[11pt]{article}

\usepackage[]{acl}
\usepackage{times}
\usepackage{latexsym}
\usepackage{todonotes}
\usepackage{booktabs}
\usepackage{multirow}
\usepackage{pifont}
\usepackage{inconsolata}
\usepackage{microtype}
\usepackage{amssymb}
\usepackage{amsmath}

\usepackage[T1]{fontenc}
\usepackage[utf8]{inputenc}

\usepackage{microtype}

\usepackage{inconsolata}

\usepackage{tikz}
\usepackage{collcell}
\usepackage{diagbox}
\usepackage{rotating}
\usepackage{pgf-pie}  

\usepackage{pgfplots}

\usepackage{amssymb}% http://ctan.org/pkg/amssymb
\usepackage{pifont}% http://ctan.org/pkg/pifont
\newcommand{\cmark}{\ding{51}}%
\newcommand{\xmark}{\ding{55}}%

\newcommand*{\MinNumber}{0.075}%
\newcommand*{\MidNumber}{0.5} %
\newcommand*{\MaxNumber}{0.978}%

\newcommand{\ApplyGradient}[1]{%
    \ifdim #1 pt > \MidNumber pt
        \pgfmathsetmacro{\PercentColor}{max(min(100.0*(#1 - \MidNumber)/(\MaxNumber-\MidNumber),100.0),0.00)} %
        \cellcolor{green!\PercentColor!yellow}{#1}
    \else
        \pgfmathsetmacro{\PercentColor}{max(min(100.0*(\MidNumber - #1)/(\MidNumber-\MinNumber),100.0),0.00)} %
        \cellcolor{red!\PercentColor!yellow}{#1}
    \fi
}

\newcolumntype{R}{>{\collectcell\ApplyGradient}c<{\endcollectcell}}
\newcommand{\highlightnew}[1]{\textcolor{black}{#1}}

\newcommand{\Sec}[1]{Section~#1}

\title{Retrieval Complexity: Measuring Question Answering Difficulty for Retrieval-Augmented Generation}

\title{Measuring Retrieval Complexity in Question Answering Systems}

\author{Matteo Gabburo$^{1}$\thanks{\ \ Work done as an intern at Amazon Alexa AI}\ , Nicolaas Paul Jedema$^{2}$\ , Siddhant Garg$^{3}$\thanks{\ \ Work completed at Amazon Alexa AI}\ , \\ \textbf{Leonardo F. R. Ribeiro$^{2}$\ , Alessandro Moschitti$^{2}$}\ \\
$^{1}$University of Trento , $^{2}$Amazon Alexa AI, $^{3}$Meta AI\\
\texttt{matteo.gabburo@unitn.it} \\ \texttt{\{jedema,leonribe,amosch\}@amazon.com} \\ \texttt{sidgarg@meta.com}
}

\begin{document}
\maketitle

\begin{abstract}
In this paper, we investigate which questions are challenging for retrieval-based Question Answering (QA). 
We (i) propose retrieval complexity (RC), a novel metric conditioned on the completeness of retrieved documents, which measures the difficulty of answering questions, and (ii) propose an unsupervised pipeline to measure RC given an arbitrary retrieval system.
Our proposed pipeline measures RC more accurately than alternative estimators, including LLMs, on six challenging QA benchmarks. 
Further investigation reveals that RC scores strongly correlate with both QA performance and expert judgment across five of the six studied benchmarks, indicating that RC is an effective measure of question difficulty.
Subsequent categorization of high-RC questions shows that they span a broad set of question shapes, including multi-hop, compositional, and temporal QA, indicating that RC scores can categorize a new subset of complex questions. 
Our system can also have a major impact on retrieval-based systems by helping to identify more challenging questions on existing datasets.
\end{abstract}

\begin{figure}[ht]
\includegraphics[width=\linewidth]{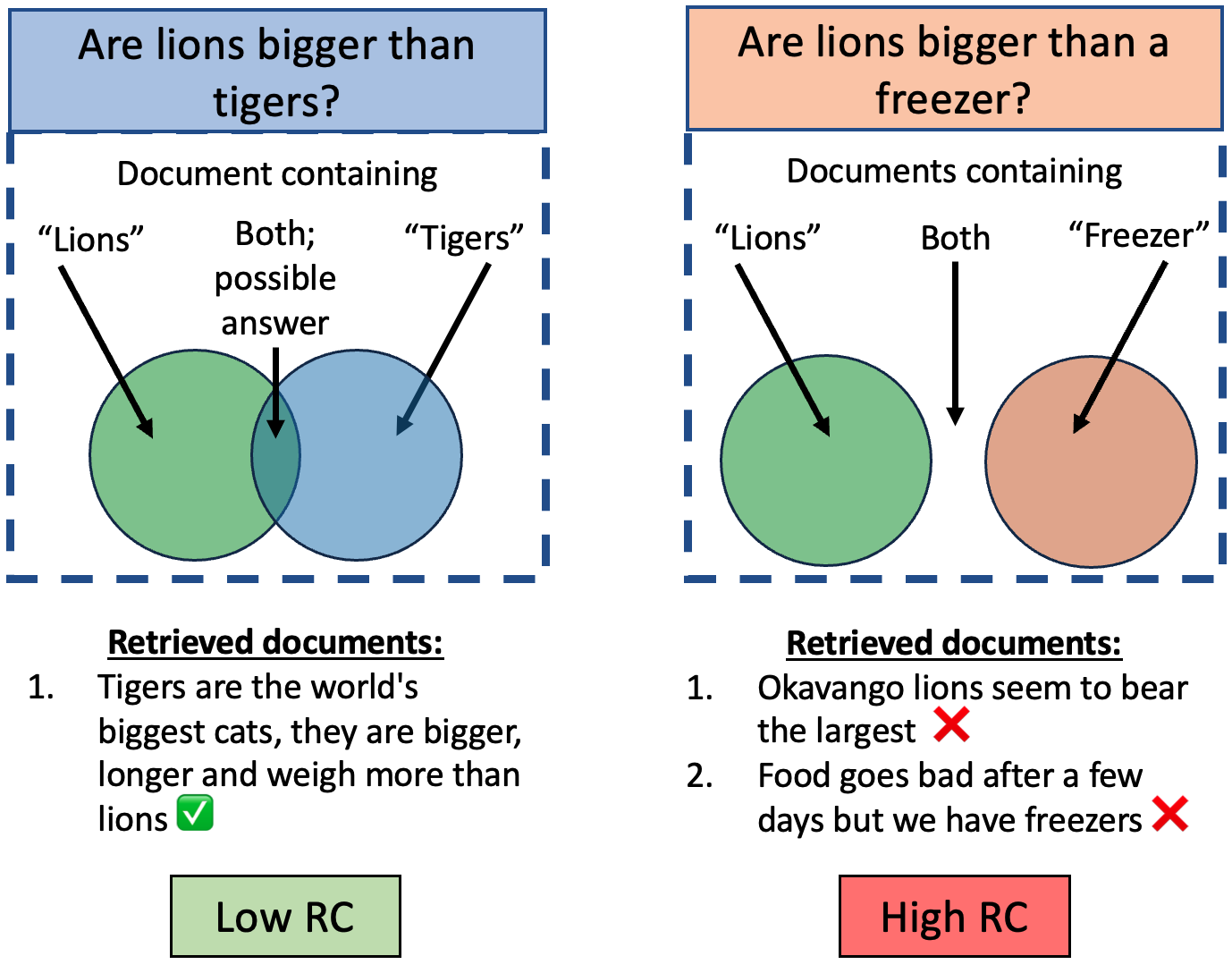}
\centering
\title{Two comparative questions with significantly different retrieval complexity}
\caption{While both questions are complex using the definition in prior works, the "Low RC" question on the left is easy for RAG systems, while the "High RC" question on the right is far more difficult because the probability of finding an existing document comparing lions and freezers is low. Documents retrieved by commercial search engines illustrate this challenge. RC correlates with the probability that the retrieved documents contain sufficient information to generate a reference answer. 
}
\label{fig:complexity_example}
\vspace{.7em}
\end{figure}

\section{Introduction}
\label{sec:introduction}

Retrieval-based approaches are a dominant paradigm in QA. They constitute a basic building for many QA pipelines, as they can gather accurate information from external sources that are useful for answering questions. Notable examples include the Retrieve and Rerank system \cite{gupta-etal-2018-retrieve,Garg_Vu_Moschitti_2020} and the recent Retrieval Augmented Generation (RAG) \cite{lewis2020rag, hsu-etal-2021-answer}.
While retrieval systems effectively address questions targeting common knowledge, they can struggle with addressing novel, domain-specific questions that require fresh, contextually accurate information. Generative models, on the other hand, bring the ability to generate responses to such challenging questions. For instance, Bing-GPT uses retrieval to answer questions about real-time events beyond the LLMs' training data cutoff \cite{dao2023performance}. 
However, retrieval augmentation can fail for challenging questions.
To address this problem, the QA community developed benchmarks to measure the performance of QA approaches on these types of questions. For example, HotPotQA \cite{yang2018hotpotqa}, ComplexWebQuestions \cite{talmor18compwebq}, and MuSiQue \cite{trivedi2022musique} challenge systems by proposing questions with "complex" shapes, such as multi-hop questions, requiring multiple steps of reasoning, e.g., "what is the GDP of the country with the tallest bridge?".
While it is assumed that these question shapes are equally difficult for RAG and other QA systems, subsequent works indicate that this is not the whole picture of QA complexity.
\highlightnew{For example, \citet{min2019compositional} show that nearly 61\% of HotPotQA's multi-hop questions actually have "single-hop answer solutions" - i.e., a single document that contains the correct answer - that a simple RAG system can use to answer correctly. More recent RAG-oriented benchmarks such as Fresh QA~\cite{vu2023freshllms} further illustrate that the complexity of the answer evidence is relevant in identifying difficult questions since the evidence can become false if not recent enough.}

In this paper, we introduce a novel metric named Retrieval Complexity and an unsupervised pipeline, named Reference-based Question Complexity Pipeline (RRCP) (\Sec\ref{sec:methodology}) to recognize it. Specifically, given a query question and one or more reference answers, RRCP employs an arbitrary retriever to procure pertinent documents. Subsequently, utilizing a novel reference-based evaluator named GenEval, RRCP scores the question based on the retrieved documents and the supplied reference answers. Conceptually, RC approximates the degree to which the evidence required to answer the question is spread across documents in the retrieval batch. The intuition behind RC is that the greater the fragmentation of required information within a retrieval batch, the harder it is for a RAG system to answer the question correctly.
For example, Fig.~\ref{fig:complexity_example} shows two questions with highly similar question semantics but significantly different retrieval quality (e.g., relevance and comprehensiveness of the information) to illustrate what RC measures.

We compare RRCP with supervised and unsupervised models, demonstrating its superior accuracy in classifying RC questions. Our unsupervised approach enables applicability to new QA benchmarks, acting as a valuable resource for the QA community.
RC is intimately linked to a specific retrieval corpus, capturing (i) index coverage, (ii) retrieval method quality, and (iii) varying semantics of questions into a single metric. To support this claim, we compute RRCP scores for six QA benchmarks using a state-of-the-art search engine (Bing) and reveal a strong correlation between RC scores and QA system performance. In addition, we show that our human expert assessment of question difficulty against retrieved evidence also correlates with RRCP scores. Finally, we present a qualitative analysis of 200 high-RC questions drawn from all benchmarks. These findings indicate a connection between high RC and diverse question types, such as multi-hop (23\%), comparative (10\%), temporal (15\%), superlative (3\%), and aggregate (16\%).

\highlightnew{In summary, our paper (i) motivates the need to estimate the answering difficulty of questions given the availability of evidence, (ii)  defines RC, a new metric that measures the difficulty of questions conditioned on a retrieval batch, (iii)  proposes RRCP, an unsupervised pipeline to measure retrieval complexity, and (iv) empirically demonstrates that our pipeline is superior at classifying high RC questions and (v) shows that higher RC scores correlate with lower QA performance.}

\section{Related work}

In this section, we introduce the related work on the main topics of our paper: question complexity, LLM-based QA, retrieval for QA, and automatic evaluation.
\label{sec:relatedwork}

\paragraph{Notions of Question Complexity:} A popular notion of question complexity in QA literature considers the number of reasoning steps (hops) required to get to the answer. 
Multi-hop questions~\cite{Mavi2022ASO} - those expected to require at least two reasoning steps - are one source of complex questions.
Popular benchmarks for multi-hop QA include HotPotQA \cite{yang2018hotpotqa} and MuSiQue \cite{trivedi2022musique}, which are constructed by aggregating multiple one-hop questions into a multi-hop one. 
\citet{min-etal-2019-compositional} highlight that such multi-hop questions may, in fact, not require multiple reasoning steps to answer, finding that many may be answered with a single snippet extracted from a relevant document. 
\citet{talmor18compwebq} suggests a related notion, compositional complexity, that assesses the degree to which the answer must be composed from multiple pieces of evidence. 
Subsequent works have shown that both multi-hop and compositional questions can be decomposed into simpler sub-questions \cite{perez2020unsupervised,yoran2023answering}, enabling a strong LLM to answer using chain-of-thought (CoT)~\cite{wei2022chainofthought}. To conclude, prior works tried to perform some performance prediction of queries \cite{querymetareview1,querymetareview2} relying on evidence in the form of different characteristic patterns in the distribution of Retrieval Status Values (RSVs).

Reasoning over temporal information - i.e., information that changes quickly over time - is another dimension of QA complexity, captured in benchmarks like Time-Sensitive-QA and FreshLLM \cite{chen2021dataset, vu2023freshllms}.
Temporal questions require systems to reflect embedded notions of temporality, which may either be explicit (e.g., Who was the president of the US in 1945?) or implicit (e.g., Who was the last president of the US?). 
Temporal knowledge graphs~\citep{saxena-etal-2021-question,shang-etal-2022-improving,sharma-etal-2023-twirgcn}, temporal embeddings in the ~\cite{huang-etal-2022-understand}, and retrieval have all been proposed as solutions to the complexity of temporal QA \cite{vu2023freshllms}.

\paragraph{LLMs for Question Answering:} Transformer-based models~\cite{vasawani2017} have become the de-facto standard for QA tasks, either in the form of encoder-only models ~\citep{Devlin2019BERTPO, Liu2019RoBERTaAR, clark2020electra}, encoder-decoder models such as T5~\citep{Raffel2020t5,lewis-etal-2020-bart} or decoder-only models \cite{radford2018improving}, most notably ChatGPT. 
Despite the impressive performance of these models across a variety of QA benchmarks, answering complex and convoluted questions without hallucinations is an enduring challenge ~\cite{huang2023survey}.
RAG systems seek to address this challenge by providing retrieved information as grounding knowledge for answer generation~\cite{lewis2020rag,gabburo-etal-2022-knowledge, borgeaud2022improving}.

\paragraph{Retrieval for QA:} BM25~\cite{crestani1999bm25survey} is a traditional retrieval approach also widely used for QA. It uses an unsupervised scoring function based on a sparse bag-of-words representation of text.
More recently, supervised retrieval systems that compute an approximate nearest neighbour score on dense representations of query and document have been proposed, such as DPR~\cite{karpukhin-etal-2020-dense} and ColBERT~\cite{khattab2020colbert}. 
Other models attempt to capture the benefits of both techniques by learning a sparse representation of documents conditioned on relevant queries, such as UniCOIL \cite{lin2021brief} and DeepImpact \cite{mallia2021learning}.

\paragraph{Automatic QA Evaluation:} Automatically evaluating the correctness of answers is a challenging task, especially in light of the compelling hallucinations created by strong generative LLMs. Previously, token-level metrics such as BLEU~\cite{papineni_bleu_2001} and sentence-level metrics such as BERTScore~\cite{sun-etal-2022-bertscore} and BLEURT~\cite{yan-etal-2023-bleurt} have been used for this task. 
However, these metrics do not correlate well with human annotation for QA tasks, as shown in \cite{gabburo-etal-2022-knowledge}. 
Recently, several works have proposed reference-based evaluation metrics for the QA and demonstrated that these have greater correlation with human annotations, such as AVA~\cite{vu-moschitti-2021-ava}, BEM~\cite{bulian-etal-2022-tomayto}, and SQuARe~\cite{gabburo2023square}.

\section{Retrieval Complexity of Questions}
\label{sec:complexquestions}

A retrieval system provides users with relevant and accurate information in response to their queries~\cite{baumgartner-etal-2022-ukp, 10.1145/3560260}. 
Recently, retrieval augmentation - that is, conditioning generation on retrieved evidence - has been shown to enable generative models to address more complex, multi-faceted questions~\cite{Luo2023ChatKBQAAG}. 
We conceptualize RC, a specific notion of question complexity, which can also be used to evaluate the complexity of questions used in prior retrieval-based QA systems, e.g., RAG models.

Intuitively, a question can be considered complex for retrieval augmentation if it cannot be answered by extracting a snippet from a retrieved document.
This dimension of complexity can occur for multiple reasons, such as if the answer requires multiple reasoning steps or a composition of supporting facts spread across retrieved content ~\cite{dua-etal-2019-drop, yang-etal-2018-hotpotqa}.
Effectively, retrieval complexity occurs when multiple pieces of information must be combined in a complex fashion to generate the correct answer, such as reasoning, comparison, or composition over multiple facts.
Multiple factors can increase the retrieval complexity, including index coverage, retrieval quality, and question syntax, as each plays a role in determining whether a document containing a snippet that answers the question is retrieved.

For example, answering the question "Are lions bigger than tigers?" (Fig.~\ref{fig:complexity_example})  requires a comparison between two related entities. 
While this question could be considered complex, the close relation of the lions and tigers increases the probability that a single snippet talks about the size of both animals. However, the probability above is an order of magnitudes lower for a similar question, such as "Are lions bigger than freezers?".

\section{Modeling Retrieval Complexity}
\label{sec:methodology}

\begin{figure}[t]
\includegraphics[width=\linewidth]{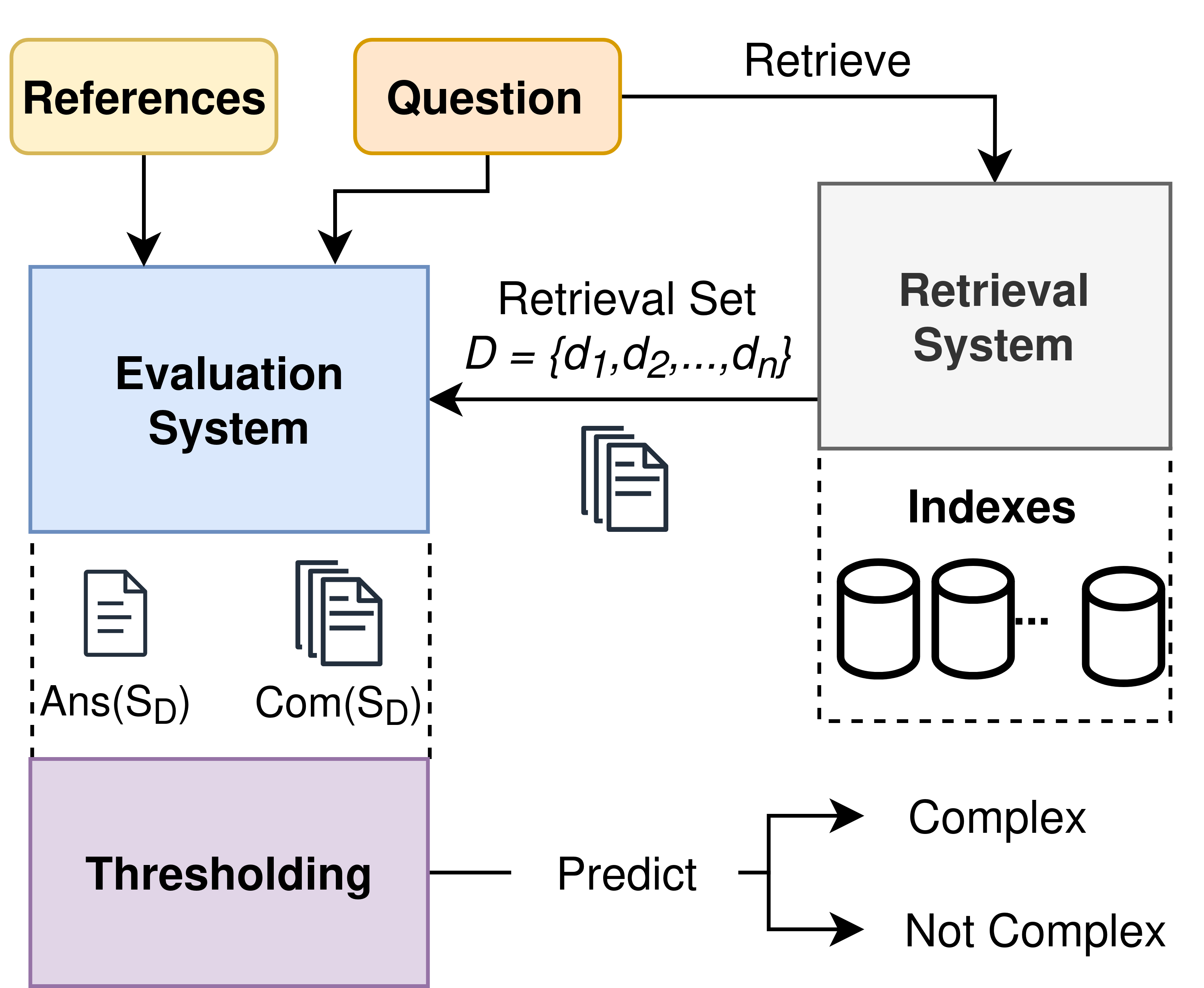}
\centering
\caption{Overview of our RRCP for measuring the complexity of questions. A retrieval system retrieves documents for the input question. These are examined by a reference-based evaluation system, which provides the probability of correctness scores. These approximate the answerability and completeness of retrieved information, i.e., an estimation of RC.}
\label{fig:pipeline}
\end{figure}

To model and recognize the retrieval complexity (\Sec\ref{sec:complexquestions}), we designed an unsupervised reference-based pipeline named RRCP (reference-based RC pipeline) shown in Fig. \ref{fig:pipeline}. RRCP is composed of three primary components: (i) A state-of-the-art retrieval system based on multiple indexes. This, given a question $Q$, composes a set of supporting documents $D=\{d_{0}, d_{1}, ..., d_{k}\}$. (ii) A powerful automatic evaluation system named GenEval, which, using the retrieved documents and a set of references,  estimates the RC of $Q$, and (iii) a constraint enforcer mechanism based on two thresholds.

\subsection{Retrieval System}
The retrieval system component of RRCP is a framework that uses multiple indexes to efficiently retrieve relevant documents in response to a given question $Q$. This system employs state-of-the-art techniques to compose a set of supporting documents, $D={d_{0}, d_{1}, ..., d_{k}}$, where each document $d_i$ is deemed relevant to the query $Q$. To ensure high reliability and robustness, we used a hybrid retrieval system based on BM25 \cite{crestani1999bm25survey} and ColBERT \cite{khattab2020colbert}.

\subsection{GenEval}
\label{sec:geneval}
Inspired by recent automatic evaluation systems such as BEM \cite{bulian-etal-2022-tomayto} and SQuArE \cite{gabburo2023square}, we designed a novel model named GenEval to recognize the difficulty of a question. GenEval is an encoder-decoder model trained to understand if a document $d_{i} \in D$ is relevant (e.g., contains a correct answer) given a question $Q$ and a set of multiple references $R=\{r_{0}, r_{1}, ..., r_{k}\}$. 

GenEval differs from BEM and SQuArE in two main aspects: 
\highlightnew{First, both SQuArE and BEM are based on an only-encoder transformer architecture \cite{Devlin2019BERTPO, Liu2019RoBERTaAR,he2021debertav3}. Using an encoder-decoder model allows for a more flexible model training starting from state-of-the-art large language models such as T5-xxl \cite{Raffel2020t5}.}

\highlightnew{Second, unlike the aforementioned approaches, GenEval has been trained on more reference datasets and on synthetic data, increasing flexibility to cases where ground-truth references are not available and using a two-headed architecture to predict the "answer correctness" and "tokens relevance". The first one estimates the probability of $d_{i}$ containing an accurate answer for $Q$, while the second computes the relevance distribution of each token of $d_{i}$ according to $Q$. RRCP uses these two inner metrics to model the Answerability and the Retrieval Completeness constraints necessary to estimate RC. We provide better details about GenEeval in Appendix \ref{apx:geneval}. }

\begin{table}[t]
\centering
\begin{tabular}{@{}lcc@{}}
\toprule
\textbf{Models} & \textbf{AE Finetuning} & \textbf{Accuracy}   \\ \midrule 
SQuARe   & \xmark & 0.572 \\
GenEval  & \xmark & \textbf{0.750} \\ \midrule
BEM      & \cmark & 0.897 \\
SQuARe   & \cmark & 0.907 \\
GenEval  & \cmark & \textbf{0.916} \\ \bottomrule
\end{tabular}
\caption{Comparison of GenEval, SQuARe and BEM on the Answer Equivalence (AE) test set. The results show that both in a zero-shot setting (without the fine-tuning on the target dataset) and in a fine-tuned setting, GenEval exhibits better performances in terms of accuracy, leading to a better correctness estimation. The better results divided per setting are in bold.}
\label{tab:genevalvsbemvssquare}
\end{table}

\highlightnew{To prove the better performance of GenEval over the BEM and SQuARe, we conducted an experiment measuring the performance of the GenEval on the AE \cite{bulian-etal-2022-tomayto} testset.}
\highlightnew{Table \ref{tab:genevalvsbemvssquare} presents the comparison results of GenEval, SQuARe, and BEM on the Answer Equivalence (AE) test set in both zero-shot and fine-tuned settings. In the zero-shot setting, without fine-tuning on the target dataset, GenEval shows better accuracy with a value of 0.750 compared to SQuARe's 0.572. In the fine-tuned setting, all models were fine-tuned on the target dataset, and the table shows that  SQuARe and GenEval exhibit improved performances, with the latter having an accuracy of 0.916, which is higher than both SQuARe's 0.907 and BEM's 0.897. These results suggest that GenEval outperforms SQuARe and BEM in terms of accuracy and correctness estimation in both settings.}

\vspace{0.3em}
\subsection{RRCP Constraints}

RRCP enforces two constraints, Answerability and Retrieval Set Completeness, to estimate RC.

\vspace{0.3em}
\subsubsection{Answerability}
\label{ssec:answerability}

From a retrieval perspective, a requirement that makes a question complex is the probability of answering it with a single document (Sec. \ref{sec:complexquestions}. \highlightnew{For example, questions like "What is the capital of France?" can be easily answered by a single document and do not fit our definition of retrieval complexity. In contrast, questions like "Is Paris bigger than the capital of the US?" have a higher degree of complexity since the probability of finding a document that makes this comparison is more difficult to find. }

We capture this property by setting a constraint on answerability: if the target question fails to meet this criterion, it is deemed potentially complex. Specifically, answerability is met if at least one document $d_{i} \in D$ contains an accurate response to $Q$. To compute the probability that an answer exists in a document, we use the probability estimated by GenEval. Specifically, we define a threshold $T_{ans} \in [0,1]$ over the probability computed by the GenEval "answer correctness" head of a document containing the correct answer $s_{i}$: if the score exceeds $T_{ans}$, the question is \emph{answerable}. Intuitively, if the question can be answered by a single retrieved document $d_{i}$ in $D$, $s_{i}$ will approach $1$.
As such, we model the answerability function $Ans(S_D)$ as follows:
\begin{equation}
Ans(S_D) = \begin{cases}
    \text{0 if } \max(s_i \in S_D) < T_{ans} \\
    \text{1 if } \exists  \, s_i \in S_D \geq T_{ans}, \label{eq:answerability} \\ 
\end{cases}
\end{equation}

\noindent where $T_{ans}$ is used to discriminate between not answerable questions when $s_{i} < T_{ans}$, and answerable questions when $s_{i} \ge T_{ans}$. 

\vspace{0.3em}
\subsubsection{Retrieval Set Completeness}
\label{ssec:completeness}
\vspace{0.3em}

\highlightnew{Retrieval set completeness determines how much the information required to answer is spread across different documents.
Indeed, a document could be partially relevant to the question but without containing sufficient information to induce the correct answer \cite{10.1145/3209978.3209980, baumgartner-etal-2022-incorporating}. For instance, to answer a question like "Who are the top 5 goalkeepers of the last ten years?" a two-year-old document could not contain all the information needed since there is no evidence about the scores achieved by the goalkeepers in the last two years. For this reason, the document is still relevant but not exhaustive. 
With the retrieval set completeness, we aim to estimate the retrieval complexity of a question by examining the heterogeneity of the retrieval set.
}
To perform this estimation, we compute the entropy of the relevance of each document in the retrieval batch $d_{i} \in D$ with each token in the question. 
For example, given the question "Are lions bigger than tigers?", we can expect that documents discussing lions will be more relevant to the question subparts that refer to lions, and the opposite for documents discussing about tigers.
Based on this notion of relevance, we can approximate whether the knowledge relevant to each portion of the question is present in the retrieval set.

To measure completeness, we leverage the relevance distribution extracted from GenEval at the token level to generate a distinct attention distribution for each document in $D$. These distributions are organized into a matrix $M$ of size $|D| \times |Q|$, where each row represents the token $t_{j}$ from the question, and columns indicate the relevance $Rel(d_i, t_j)$ of each document $d_{i}$ to those tokens. 

Finally, to ensure comparability, we normalize the entropy for each document and compute the average normalized entropy to obtain the completeness score $S_{D}$:
\begin{equation}
	\begin{split}
	S_{D} &= \frac{ \sum\nolimits_{i=1}^{|D|} \left\lVert \sum\nolimits_{j=1}^{|Q|}  Rel(d_i, t_j) \right  \lVert}{|D|} \, .
	\end{split}
\end{equation}

Similar to answerability, we apply a threshold $T_{com}$ to the completeness score:
\begin{equation}
Com(S_D) = \begin{cases}
\text{1 if } S_{D} \geq T_{com} \label{eq:completeness} \\
\text{0 if } S_{D} < T_{com}
\end{cases}
\end{equation}

\vspace{0.3em}
\subsection{Classifying Complex Questions}
While answerability and retrieval set completeness are two standalone criteria, RRCP leverages both signals to approximate retrieval complexity (RC). 
Specifically, RRCP considers a question complex when the question is not answerable with a single document and when the retrieval set is incomplete. 
Formally, we consider both \ref{eq:answerability} and \ref{eq:completeness}, classifying a question as retrieval-complex if $Ans(S_D) = 0$ and $Com(S_D) = 1$. 
In addition to enabling RC classification, answerability and completeness scores can be used as additional diagnostic metrics.
 
\section{Experiments}
\label{sec:experiments}

\highlightnew{In this section, we focus on studying Retrieval Complexity (RC) and our RRCP framework for its detection. First, in sections \ref{ssec:datasets} and \ref{ssec:RRCP_setting} we describe the datasets we considered and the setup of RRCP. Then, we validate RC and RRCP by conducting both quantitative and qualitative analyses. Specifically, we first compare RRCP against a strong prompted LLM unsupervised baseline directly on a set of $4$ selected datasets in, targeting the specific complexity class for each of them (\Sec\ref{ssec:cq_identification}). Secondly, in \Sec\ref{ssec:answer_correctness_generative}, we evaluate the correlation between the RC classification (complex/not complex) with LLM answer capability. This allows us to understand some possible limitations of our notion of complexity. In the third quantitative experiment (\Sec\ref{ssec:answer_correctness_search_engines}), we measure the answerability of the questions identified as complex by our pipeline using a state-of-the-art search engine, such as Bing. Finally, we perform a qualitative analysis to inspect the limitations and the generalizability of our approach (\Sec\ref{sec:qualitative_analysis}).}

\subsection{Datasets}
\label{ssec:datasets}
We evaluated RRCP on the following academic benchmarks: ComplexWebQuestions (CWQ) \cite{talmor18compwebq}, HotPotQA \cite{yang2018hotpotqa}, StrategyQA \cite{geva2021strategyqa}, and MuSiQue \cite{trivedi2022musique}.
We used the question complexity information provided in each dataset to define "complex" or "not complex" labels. 
For CWQ, we labelled their simple questions as not complex and the more challenging ones composed from them as complex.
For HotPotQA, we used the "difficulty level" associated with each question in the dataset. 
For MuSiQue and StrategyQA, we considered one-hop questions as not complex and multi-hop ones as complex. 

We also used Natural Questions \cite{kwiatkowski2019nq} and QuoraQP-a \cite{wang2020match} to evaluate our pipeline on more natural user-generated questions. 
A small team of expert annotators determined complexity labels on these datasets using the instructions outlined in Appendix \ref{apx:manualannotation}.
Table \ref{tab:dataset_complexity_classes} reports the complexity categories we found in some of the datasets above. Additional details regarding the distribution, size, and splits of these datasets are defined in Appendix \ref{apx:datasets}.

\input{tables/datasets_complexity_classes}

\vspace{0.3em}
\subsection{RRCP setup}
\label{ssec:RRCP_setting}
\vspace{0.3em}

We implemented RRCP (described in \Sec\ref{sec:complexquestions}), with a state-of-the-art hybrid retrieval system based on BM25 \cite{crestani1999bm25survey} and ColBERT \cite{khattab2020colbert} with an index of documents containing (Wikipedia~\cite{petroni-etal-2021-kilt}, and MS MARCO~\cite{nguyen2016ms}).

To implement the automatic evaluation system represented by GenEval (Sec.\ref{sec:geneval}), we trained a T5-xxl model \cite{Raffel2020t5} on two existing datasets, which are WQA \cite{vu-moschitti-2021-ava} and AE \cite{bulian-etal-2022-tomayto}. We discuss additional details regarding the experimental setting and the performance against other automatic evaluation metrics in Appendix \ref{apx:geneval}. We set thresholds $T_{ans}=0.15$ and $T_{com}=0.80$ based on a small set of $50$ manually written test questions. 

\vspace{0.3em}
\subsection{Complex Question Identification}
\label{ssec:cq_identification}
\vspace{0.3em}

To compare the ability of RRCP to detect complex questions with alternative approaches (e.g., LLMs, supervised models), we use the answerability and completeness scores to classify questions as described in \Sec\ref{ssec:cq_identification}.  
We compare the performance in terms of accuracy and f1-measure against a strong unsupervised baseline consisting of a prompted state-of-the-art LLM for this task.
We report the results of these experiments in Table~\ref{tab:complex_questions_classification}. 

\input{tables/automatic_evaluation_pipeline_acc}

Furthermore, we ablated different configurations of the pipeline to assess the benefits given by the two constraints.
Combining the answerability and the retrieval set completeness constraints proved to be beneficial, enabling a more accurate classification than the alternatives.
RRCP, based only on the answerability constraint, generally demonstrates higher results than the LLM baseline. Also, without considering the completeness constraint, the pipeline provides accurate predictions. 
On the other hand, the model only based on retrieval completeness achieves lower performance than the other approaches, highlighting the fact that it can not be used standalone. 

\subsection{LLM Performance on Complex Questions}
\label{ssec:answer_correctness_generative}

In this section, we study whether RRCP predictions (complex vs. not complex) correlate with the notion of complexity of LLM-based systems (only using parametric knowledge).
Specifically, we evaluate the correlation between its RC classification and LLM answer capacity (0/1 label), where the latter is computed by (i) generating an answer with LLM and (ii) manually annotating its correctness.  

For each dataset in \Sec\ref{ssec:datasets}, we apply RRCP, employing our defined criteria to filter complex from non-complex questions. We use an end-to-end QA system composed of a prompted Mistral 7B \cite{Jiang2023Mistral7} (a large language model) designed to receive input questions and generate accurate answers. Then, we evaluated each generated answer, comparing it with the original gold answers in the datasets. Finally, we measured the agreement between the detected RC and answer correctness in terms of Pearson Correlation. 

We present the results of this analysis in terms of accuracy in Table~\ref{tab:accuracy_generated_answers}, where we show the accuracy scores for both complex and non-complex questions across the different datasets and their Pearson Correlation. Each entry in the table represents the percentage of questions correctly answered by this LLM-based QA system.

\begin{table}[h]
\small
\resizebox{\columnwidth}{!}{
\begin{tabular}{lccc}
\toprule
\textbf{Dataset} & \textbf{Complex} & \textbf{Not Complex} & \textbf{PCC} \\
\midrule
CWQ & 0.328 & 0.641 & \multirow{6}{*}{0.449}  \\
HotPotQA & 0.266 & 0.328 & \\
StrategyQA & 0.000 & 0.078 & \\
MuSiQue & 0.031 & 0.563 & \\
Natural Questions & 0.023 & 0.547 & \\
Quora & 0.016 & 0.016 & \\
\bottomrule
\end{tabular}}
\caption{Average answerability of answers generated for complex and not-complex questions selected by our pipeline. The results show that the questions marked as complex are generally more difficult to answer. PCC denotes the correlation in terms of the Pearson Correlation Coefficient between the complex and not complex questions in terms of answerability.}
\label{tab:accuracy_generated_answers}
\vspace{0.3em}
\end{table}

The results confirm that our approach effectively identifies complex questions, resulting in higher answerability. Questions identified as complex by the pipeline generally exhibit higher accuracy in terms of answerability, indicating that these questions are easier to be answered for the model. Specifically, questions categorized as complex in CWQ, HotPotQA, MuSiQue, and Natural Questions display higher accuracy than those classified as not complex. Notably, the complexity assessment in Quora and StrategyQA does not significantly impact the question answerability, suggesting a different nature of complexity in this dataset or some bias. In StrategyQA, complex questions were labelled due to limited "true/false" reference answers, impacting evaluation metrics. Modifying the prompt eliminated this artifact, highlighting the need for precise evaluation metrics. In contrast, Quora's complexity designation was influenced by poor-quality reference answers, emphasizing caution when interpreting complexity labels in datasets with sub-optimal reference quality.

\subsection{Search Engine Performance on Complex Questions}
\label{ssec:answer_correctness_search_engines}

We further explore the quality of our pipeline by sampling a set of 500 questions from the "complex" questions of each dataset and analyzing how many of them can be answered by a state-of-the-art search engine such as Bing\footnote{\url{https://www.bing.com/}}. This dataset is constructed by selecting 256 questions from datasets with their own notion of complexity, including CWQ, HotPotQA, StrategyQA, and MuSiQue (64 questions per dataset), and another 256 from natural datasets not specifically focused on complexity (128 each from Natural Questions and Quora).
To conduct the evaluation, we select a pool of expert annotators. Their task is to determine whether a given question can be answered by examining the top 5 search engine results. If the answer is found within these results or can be inferred through reasoning based on them, the question is considered answerable. Furthermore, the annotators are tasked with evaluating whether the question aligns with our predefined notion of retrieval complexity as described in \Sec\ref{sec:complexquestions}.

\begin{table}[t]
\resizebox{\columnwidth}{!}{
\begin{tabular}{c|cc|c}
\toprule
\textbf{RRCP} & \multirow{2}{*}{\textbf{Answerable}} & \textbf{Not} &  \multirow{2}{*}{\textbf{MCC}}\\
\textbf{Classification} & & \textbf{Answerable} & \\
\midrule
\textbf{Not Complex} & 0.600 & 0.113 &  \multirow{2}{*}{0.508} \\
\textbf{Complex} & 0.400 & 0.887 & \\
\bottomrule
\end{tabular}}
\caption{Probability for a question classified by RRCP of being answered by a state-of-the-art search engine. The answerability is computed using manual annotators. MCC stands for Matthew Correlation Coefficient and measures the correlation between the complexity classification and its answerability.}
\label{tab:search_engine}
\vspace{0.3em}
\end{table}

The results of our analysis, detailed in Table \ref{tab:search_engine}, reveal significant insights. Indeed, the probability of a question marked as complex by RRCP of not being answered by a state-of-the-art search engine is high ($0.887$). Similarly, a question marked as not complex has a high chance to be answered ($0.6$). In addition, the results show a good correlation between question-predicted complexity and their answerability (manually annotated), with a Matthew correlation of $0.508$, indicating a moderate to strong positive correlation between the predicted classifications and the actual classifications.

\subsection{Limitations of Supervised Approaches}
\label{ssec:supervised_baseline}

In this section, we explore the limitations of employing supervised approaches to estimate question complexity. To show this finding, we trained several cross-encoder models on the datasets introduced in \Sec\ref{ssec:datasets} using a supervised learning approach. The experimental setup is detailed in Appendix \ref{apx:supervisedparams}. The outcomes of this analysis are presented in Table \ref{tab:supervised_results}.

\input{tables/supervised_results}

We assessed the resulting five models on the four datasets. Notably, Natural Questions \cite{kwiatkowski2019nq} and QuoraQP-a \cite{wang2020match} lack an internal definition of complexity (Table \ref{tab:dataset_complexity_classes}), making automatic evaluation and training impracticable.

Although the model trained on the merged dataset demonstrated impressive performance, individual models struggled to replicate this strong performance across datasets. This disparity implies that fine-tuning induces models to overfit on the question distribution rather than learning the intricacies of question complexity. This observation highlights the need for standard fine-tuning to develop dataset-agnostic models to detect question complexity.
It is important to acknowledge that the "ALL" model performs well due to its ability to identify which dataset the test question belongs to, thanks to the distinctive topics and typical question shapes characterizing each dataset. Consequently, supervised models only capture specific properties of the data and complexity definitions.

\subsection{Qualitative Analysis and Limitations}
\label{sec:qualitative_analysis}

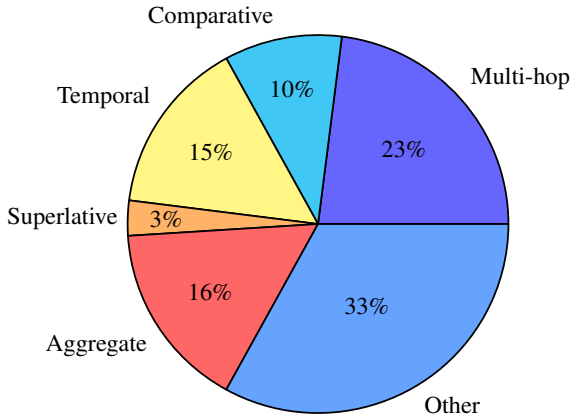
\begin{figure}
    \centering
    \resizebox{1\columnwidth}{!}{
    \begin{tikzpicture}
    \pie{23/Multi-hop,
    	10/Comparative,
    	15/Temporal,
    	3/Superlative,
    	16/Aggregate,
    	33/Other}
    \end{tikzpicture}}
    \caption{Distribution of complexity classes from a sample of 200 high RC questions selected among the datasets described in \Sec\ref{ssec:datasets}.}
    \label{fig:complexity_distribution}
\vspace{0.3em}
\end{figure}

We further examine the behaviour of our pipeline through a rigorous qualitative evaluation. We notice that RRCP, initially designed to recognize retrieval complexity, can address other complexity classes beyond its primary scope. These complexity classes are part of a wide range of question types, such as comparative questions (e.g., "Is an elephant bigger than a cat?"), multi-hop questions (e.g., "How old is the wife of the tallest NBA player?"), questions needing the "aggregation" of multiple answers in the form of disconnected entities (e.g., "List every football player who played in the last World Championship?"), time-based questions, both implicit and explicit (e.g., "Who is the current president of the US?" and "Which movie won the Oscar for the best movie in 1992?") and superlative questions (e.g., "What are the best countries to travel to in March?") following the distribution shown in Fig.~\ref{fig:complexity_distribution}. 

However, our analysis also highlights notable limitations within our pipeline. We identified two primary challenges. Firstly, the reference-based nature of our approach is susceptible to the quality of the references used. Consequently, the predictive accuracy of the pipeline is significantly impacted by the quality of these references. This issue becomes evident during the examination of results from Quora, where we observed a discernible drop in performance. Despite employing a robust retrieval system, the precision of our pipeline is closely linked to the quality of the references. This finding underscores the critical role of reference quality in shaping the effectiveness of our methodology. Addressing this challenge necessitates a comprehensive evaluation of the reference sources employed and potential enhancements in the retrieval system to mitigate the adverse effects on prediction accuracy. By acknowledging and addressing these limitations, our research aims to refine the qualitative analysis pipeline, ensuring its applicability across various complexity classes and enhancing its overall robustness in handling diverse question types.

\section{Retrieval Complexity Applications}
\label{sec:applications}
In Section \ref{sec:experiments}, we show that RRCP is a valuable tool which allows for the recognition of RC with high accuracy. This could be difficult to achieve with a reference-free approach. Although the need of references could limit the applicability of RRCP in real-world scenarios, it serves as a crucial step towards understanding the potential applications and future directions of RC in the field of QA systems. Indeed, RC is a valuable diagnostic tool that can be applied to various settings, including question routing, optimizing document usage in Relevance and Ranking (RAG) systems, filtering QA datasets, and identifying interesting questions according to some complexity notion. The motivating concepts of RC are potentially applicable to many settings, as they provide insights into the underlying complexity of questions and can help reveal challenging questions in existing academic benchmarks.

\section{Conclusion}
\label{sec:conclusion}

\highlightnew{In this paper, we introduced a novel concept of question complexity, measuring the difficulty of finding accurate answers from multiple sources. By combining a top-tier retrieval system with an effective automatic evaluation system, our approach (RRCP) demonstrated high accuracy in handling complex questions that go beyond single-source responses. We performed an extensive experimentation, conducting both quantitative and qualitative analyses on various datasets, proving the robustness of our pipeline in managing and recognizing complex question types, including multi-hop questions, time-based queries, and comparative inquiries. The results consistently showed that our approach outperformed state-of-the-art supervised transformer models and large language models in these challenging scenarios. Additionally, our benchmark evaluations underscored the superiority of our method, highlighting its capability to deliver more precise answers where traditional models struggled. The empirical evidence confirmed that our pipeline is particularly effective in classifying high retrieval complexity (RC) questions and that higher RC scores were correlated with lower QA performance, underscoring the validity of our RC metric.}
\highlightnew{As future work, we plan to reduce the reference dependency of the Retrieval Complexity Pipeline (RRCP) in our evaluation system by incorporating large language models (LLMs) and parametric knowledge. This enhancement aims to increase the number of application scenarios for our framework, broadening its usability and effectiveness in diverse contexts. By leveraging these advanced models, we anticipate further improvements in the efficiency and accuracy of our system, making it even more versatile in handling a wide range of complex queries.}

\section{Limitations}
\label{sec:limitations}

While our study provides valuable insights about retrieval complexity and our framework proved to be a useful tool to determine whether a question fits with our notion of complexity, it is important to acknowledge several limitations that may impact our approach:

\paragraph{Corpus dependency:} As mentioned in \Sec\ref{sec:methodology} and \Sec\ref{sec:experiments}, the performance of our pipeline is limited by the goodness of the corpora considered to build the index. For example, this limitation is evident when thinking about a time-based question asking about recent events not covered by our index. We partially mitigate this problem by working on the completeness constraints as specified in \Sec\ref{sec:methodology}, but we consider this limitation as an open problem, planning to address it in future work.

\paragraph{Reference-based approach:} The core of our framework is based on a reference-based evaluation system. Although we can rely on high accuracy and human-like performance in terms of correctness evaluation, on the other hand, we are limited by the need for gold answers to get a good answerability prediction, limiting the scope of RRCP on labeled data. However, despite this limitation, we believe that larger reference-free approaches based on LLMs could soon replace the automatic evaluation component. 

\paragraph{Quality of the retrieval system:} Our methodology accuracy scales with the quality and accuracy of the retrieval system. 

\paragraph{Thresholds:} Determining $T_{ans}$ and $T_{com}$ is a trivial task that can be done with few examples. However, their values are critical to determine the RC with high accuracy.
\\
\\
As mentioned in \Sec\ref{sec:experiments} and \Sec\ref{sec:conclusion}, we plan to address these limitations in future work.

\bibliography{anthology,custom}

\appendix

\section{Appendix}
%\label{sec:appendix}

\subsection{Modelling Details}
\label{apx:supervisedparams}

The supervised model was trained on Amazon AWS P3dn.24xlarge hosts using specific hyperparameters selected after a parameter search. The model architecture utilized the roberta-base \cite{su-etal-2022-roberta} configuration, with a batch size (bs) of $256$ instances. We used an Adam optimizer considering a learning rate (lr) of $1e-05$ during training, carried out over $10$ epochs. The model selection criterion was based on achieving the highest F1 score on the development set (devset), ensuring the selection of the most effective model variant. Additionally, the training process incorporated mixed-precision arithmetic (fp16) to enhance computational efficiency and speed up the training procedure. We estimate that GPU hours used for baseline training and pipeline inference to be no more than 462 GPU hours.

\subsection{Datasets}
\label{apx:datasets}

In this section, we provide an exhaustive description of the datasets we considered in the paper to validate our approach. 

\paragraph{ComplexWebQuestions} \cite{talmor18compwebq} is a dataset for answering complex questions that require reasoning over multiple web snippets. It contains 34686 examples in total (27368, 3518, and 3530 for train, dev and test splits), and each example presents a question, an answer, and a SPARQL query to retrieve the web snippets needed to build the context.

\paragraph{HOTPOTQA} \cite{yang2018hotpotqa}: A QA dataset designed to contain complex questions that can not be answered without reasoning and additional context. To support this, they also added different paragraphs for each question in the dataset that can be used as a context to provide the answer. 

\paragraph{StrategyQA} \cite{geva2021strategyqa}: StrategyQA is a question-answering benchmark focusing on open-domain questions where the required reasoning steps are implicit in the question and should be inferred using a strategy. StrategyQA includes 2,780 examples, each consisting of a strategy question, its decomposition, and evidence paragraphs. The only usable split is the "train "since it has both the decompositions and the facts.

\paragraph{MuSiQue} \cite{trivedi2022musique}: This dataset has been prepared by joining multi-hop questions with single-hop questions from different datasets using a bottom-up approach. Differently from BREAK, the decomposed questions here look more natural. 

\paragraph{Natural Questions} \cite{kwiatkowski2019nq}: Large scale dataset made considering real Google queries. Each query is paired with a corresponding Wikipedia page and the relevant passage containing the answer. By definition, this dataset does not contain questions that fit our definition of complexity (\Sec\ref{sec:complexquestions}) since, for construction, the majority of the questions have answers that can be answered by a single passage. However, studying this dataset is helpful to recognize the limitations of the retrieval system used by the pipeline and to study the correlations between what is complex for a human and for a QA system.

\paragraph{QuoraQP-a} \cite{wang2020match}, is a question-answering dataset made pairing existing questions from Quora Question Pairs (QQP) with their original answers. 
    
\subsection{Supervised results}

For completeness, in Table \ref{tab:supervised_ablation}, we report the complete results obtained by the supervised approach measured in terms of Accuracy, Precision, Recall, and F1.

\input{tables/supervised_results_apx}

\subsection{GenEval details}
\label{apx:geneval}

The GenEval is an encoder-decoder transformer model based on the T5-XXL \cite{Raffel2020t5}. We trained the model on three datasets, WQA \cite{gabburo2023square}, and AE \cite{bulian-etal-2022-tomayto} and ASNQ \cite{Garg_Vu_Moschitti_2020}. Differently from the original SQuArE, we used a variable number of references during the training, considering also generated ones (we generated these references using a GenQA model \cite{gabburo-etal-2022-knowledge}).
We experimented with different parameters, and we found the best combination of parameters training the model for $20$ epochs on every dataset using a batch size of $32$, $fp32$, and Adam as optimizer with a learning rate equal to $5e-05$. We select the best checkpoint by evaluating the AUROC (Area Under the Curve) on the validation set.

\subsection{Mistral 7B Evaluation Prompt}

To perform the automatic evaluation on LLM, we are considering the following prompt:
\\
\textit{<s>[INST]"Consider the following question Q: \{s\}. Is question Q complex according to the provided definition? A complex question cannot be answered by a single document; it necessitates reasoning over different snippets due to the low probability of finding the answer within existing sources. Examples of complex questions include inquiries like 'Is a cup of tea bigger than an elephant?' where the comparison is unlikely to be found in a single document. In contrast, questions such as 'Is an elephant bigger than a lion?' are not complex because 'elephant and lions' can be part of the same document with high probability. Please respond with 'yes' if the question is complex and 'no' if it is not. Ensure your reply is concise, strictly limited to 'yes' or 'no'." [/INST]}

\subsection{Manual annotation}
\label{apx:manualannotation}

To perform the manual annotation described in Section \ref{ssec:answer_correctness_search_engines}, we employed a set of $5$ voluntary international experts (from the US, Brazil, India, and Italy) in the QA domain to annotate the data. We instruct each of them on the task, providing the notion of retrieval complexity \ref{sec:complexquestions} and a set of examples to reference during the annotation.
Specifically, for each question present in our evaluation set, the annotation has been done by presenting the question to the annotator, the $top-5$ web pages retrieved by the search engine, and the top document retrieved using our retrieval system. Then, for each question, the annotators determine whether the question is answerable by one or more Bing search results (applying or inferring reasoning if needed). During the annotation process, the annotator was unaware of the RC assigned by RRCP.

\end{document}

%% file: tables/datasets_complexity_classes.tex
\begin{table}[]
\resizebox{1\linewidth}{!}{
\centering
\begin{tabular}{@{}lcc@{}}
\toprule
\multirow{2}{*}{\textbf{Dataset}}  & \multicolumn{2}{c}{\textbf{Complexity Class}} \\ 
& Compositional &  Multihop\\ \midrule
CWQ                  \cite{talmor18compwebq} &  \cmark   &  \cmark   \\
HotPotQA             \cite{yang2018hotpotqa} &  \xmark                    &  \cmark   \\
StrategyQA           \cite{geva2021strategyqa} &  \cmark    &  \cmark   \\
MuSiQUe              \cite{trivedi2022musique} &  \xmark                  &  \cmark   \\

Natural Questions    \cite{kwiatkowski2019nq} &  \xmark                  &  \xmark \\
QuoraQP-a            \cite{wang2020match} &  \xmark                  &  \xmark 

\\ \bottomrule
\end{tabular}
}
\vspace{.3em}
\caption{Categorization of datasets into compositional and multihop complexity classes based on their respective characteristics.}
\vspace{.3em}
\label{tab:dataset_complexity_classes}
\end{table}

%% file: tables/automatic_evaluation_pipeline_acc.tex
\begin{table*}[ht]
\resizebox{1\linewidth}{!}{
\begin{tabular}{@{}lccccccccccc@{}}
\toprule
\multirow{2}{*}{\textbf{Approach}} & \multirow{2}{*}{\textbf{Answerability}} & \textbf{Retrieval Set} & \multicolumn{2}{c}{\textbf{CWQ}} & \multicolumn{2}{c}{\textbf{HotPotQA}} & \multicolumn{2}{c}{\textbf{StrategyQA}} & \multicolumn{2}{c}{\textbf{MuSiQue}} & \textbf{Average} \\
                          & & \textbf{Completeness} &   ACC & F1    & ACC   & F1    & ACC   & F1    & ACC    & F1 & F1 \\ \midrule
%Supervised                &\xmark&\xmark& 0.994 & 0.974 & 0.799 & 0.834 & 0.941 & 0.956 &  0.944 & 0.956 & 0.930 \\
LLM                       &\xmark&\xmark& 0.649 & 0.780 & 0.699 & 0.815 & 0.548 & 0.436 &  0.582 & 0.709 & 0.685 \\ \midrule
\multirow{3}{*}{RRCP} & \cmark & \xmark & 0.905 & 0.872 & 0.611 & 0.731 & \textbf{0.754} & 0.737 &  \textbf{0.799} & 0.834 & 0.794 \\
                          & \xmark & \cmark & 0.818 & 0.189 & 0.560 & 0.674 & 0.554 & 0.493 &  0.679 & 0.651 & 0.502 \\ 
                          & \cmark & \cmark & \textbf{0.912} & \textbf{0.882} & \textbf{0.718} & \textbf{0.829} & 0.623 & \textbf{0.742} &  0.780 & \textbf{0.838} & \textbf{0.823} \\
\bottomrule
\end{tabular}}
\caption{Performance comparison of our pipeline against a complex question classifier in terms of Accuracy and F1 (the best results are in bold). The pipeline has been tested ablating the two constraints (Answerability and Retrieval Completeness) and on different Complex QA datasets. The results show that our pipeline is able to identify complex questions with high accuracy and that combining the two constraints is beneficial. For these experiments, we selected the best pipeline thresholds based on the development set.}
 \label{tab:complex_questions_classification}
\end{table*}

%% file: tables/supervised_results.tex
\begin{table}[ht]
\centering
\resizebox{1\linewidth}{!}{
\begin{tabular}{cl|cccc}
      \toprule
      & \multirow{2}{*}{\textbf{Datasets}} & \multicolumn{4}{c}{\textbf{Evaluated}} \\
                 & & CWQ   & HotPotQA & MuSiQue & StrategyQA \\ 
                 \midrule
      \multirow{5}{*}{\rotatebox[origin=c]{90}{\textbf{Trained}}}
      & ALL        & 0.974 & 0.834 & \textbf{0.956} & 0.921 \\ 
      & CWQ        & \textbf{1.000} & 0.448 & 0.609 & 0.266 \\
      & HotPotQA   & 0.901 & \textbf{0.842} & 0.714 & 0.656 \\
      & MuSiQue    & 0.851 & 0.491 & 0.861 & 0.618 \\
      & StrategyQA & 0.075 & 0.253 & 0.238 & \textbf{0.978} \\
      \bottomrule
\end{tabular}
}
\vspace{.3em}
\caption{The supervised approach displays limited effectiveness for cross-dataset evaluation, indicating reduced performance when trained on one dataset and tested on another. Despite similar complexity classes between datasets (e.g., Multihop), the transferability of models across distinct datasets (e.g., training on CWQ and testing on HotPotQA) exhibits notable performance discrepancies. Notice that ALL refers to a model trained on all the datasets combined together.}
\label{tab:supervised_results}
\end{table}

%% file: tables/supervised_results_apx.tex
\begin{table}[t]
\centering
\resizebox{1\linewidth}{!}{
\begin{tabular}{@{}lccccc@{}}
\toprule
\textbf{Datasets} & \multicolumn{1}{c}{\textbf{Accuracy}} & \multicolumn{1}{c}{\textbf{Precision}} & \multicolumn{1}{c}{\textbf{Recall}} & \multicolumn{1}{c}{\textbf{F1Score}} \\ \midrule 
\multicolumn{5}{c}{\textbf{Trained on All}} \\ \midrule 
CWQ &0.994 & 0.956 & 0.994 & 0.974 \\
HotPotQA & 0.799 & 0.891 & 0.799 & 0.834 \\
MuSiQue & 0.944 & 0.970 & 0.944 & 0.956 \\
StrategyQA & 0.941 & 0.913 & 0.941 & 0.921 \\ \midrule
\multicolumn{5}{c}{\textbf{Trained on CWQ}} \\ \midrule 
CWQ & 1 & 1 & 1 & 1 \\
HotPotQA & 0.501 & 0.641& 0.501& 0.448\\
MuSiQue & 0.619& 0.858& 0.619& 0.609\\
StrategyQA & 0.501& 0.680& 0.501& 0.266\\ \midrule
\multicolumn{5}{c}{\textbf{Trained on HOTPOTQA}} \\ \midrule
CWQ & 0.905& 0.896 & 0.905 & 0.901\\
HotPotQA & 0.808 & 0.898 & 0.808 &0.842 \\
MuSiQue & 0.697& 0.873& 0.697& 0.714\\
StrategyQA & 0.682& 0.668& 0.682& 0.656\\ \midrule
\multicolumn{5}{c}{\textbf{Trained on MUSIQUE}} \\ \midrule 
CWQ & 0.932& 0.802& 0.932& 0.851\\
HotPotQA & 0.507& 0.519& 0.507& 0.491\\
MuSiQue & 0.835 & 0.926& 0.835 & 0.861\\
StrategyQA & 0.647& 0.637& 0.647& 0.618\\ \midrule
\multicolumn{5}{c}{\textbf{Trained on StrategyQA}} \\ \midrule 
CWQ & 0.388& 0.175& 0.388& 0.075\\
HotPotQA & 0.512& 0.523& 0.512& 0.253\\
MuSiQue & 0.435& 0.214& 0.435& 0.238 \\
StrategyQA & 0.984& 0.974& 0.984& 0.978\\ \bottomrule
\end{tabular}
}
\caption{Results of the supervised baseline on the different settings and measured in terms of accuracy, precision, recall and F1Score}
\label{tab:supervised_ablation}
\end{table}

%% file: main.bbl
\begin{thebibliography}{58}
\expandafter\ifx\csname natexlab\endcsname\relax\def\natexlab#1{#1}\fi

\bibitem[{Baumg{\"a}rtner et~al.(2022{\natexlab{a}})Baumg{\"a}rtner, Ribeiro, Reimers, and Gurevych}]{baumgartner-etal-2022-incorporating}
Tim Baumg{\"a}rtner, Leonardo F.~R. Ribeiro, Nils Reimers, and Iryna Gurevych. 2022{\natexlab{a}}.
\newblock \href {https://doi.org/10.18653/v1/2022.emnlp-main.614} {Incorporating relevance feedback for information-seeking retrieval using few-shot document re-ranking}.
\newblock In \emph{Proceedings of the 2022 Conference on Empirical Methods in Natural Language Processing}, pages 8988--9005, Abu Dhabi, United Arab Emirates. Association for Computational Linguistics.

\bibitem[{Baumg{\"a}rtner et~al.(2022{\natexlab{b}})Baumg{\"a}rtner, Wang, Sachdeva, Geigle, Eichler, Poth, Sterz, Puerto, Ribeiro, Pfeiffer, Reimers, {\c{S}}ahin, and Gurevych}]{baumgartner-etal-2022-ukp}
Tim Baumg{\"a}rtner, Kexin Wang, Rachneet Sachdeva, Gregor Geigle, Max Eichler, Clifton Poth, Hannah Sterz, Haritz Puerto, Leonardo F.~R. Ribeiro, Jonas Pfeiffer, Nils Reimers, G{\"o}zde {\c{S}}ahin, and Iryna Gurevych. 2022{\natexlab{b}}.
\newblock \href {https://doi.org/10.18653/v1/2022.acl-demo.2} {{UKP}-{SQUARE}: An online platform for question answering research}.
\newblock In \emph{Proceedings of the 60th Annual Meeting of the Association for Computational Linguistics: System Demonstrations}, pages 9--22, Dublin, Ireland. Association for Computational Linguistics.

\bibitem[{Borgeaud et~al.(2022)Borgeaud, Mensch, Hoffmann, Cai, Rutherford, Millican, van~den Driessche, Lespiau, Damoc, Clark, de~Las~Casas, Guy, Menick, Ring, Hennigan, Huang, Maggiore, Jones, Cassirer, Brock, Paganini, Irving, Vinyals, Osindero, Simonyan, Rae, Elsen, and Sifre}]{borgeaud2022improving}
Sebastian Borgeaud, Arthur Mensch, Jordan Hoffmann, Trevor Cai, Eliza Rutherford, Katie Millican, George van~den Driessche, Jean-Baptiste Lespiau, Bogdan Damoc, Aidan Clark, Diego de~Las~Casas, Aurelia Guy, Jacob Menick, Roman Ring, Tom Hennigan, Saffron Huang, Loren Maggiore, Chris Jones, Albin Cassirer, Andy Brock, Michela Paganini, Geoffrey Irving, Oriol Vinyals, Simon Osindero, Karen Simonyan, Jack~W. Rae, Erich Elsen, and Laurent Sifre. 2022.
\newblock \href {http://arxiv.org/abs/2112.04426} {Improving language models by retrieving from trillions of tokens}.

\bibitem[{Bulian et~al.(2022)Bulian, Buck, Gajewski, B{\"o}rschinger, and Schuster}]{bulian-etal-2022-tomayto}
Jannis Bulian, Christian Buck, Wojciech Gajewski, Benjamin B{\"o}rschinger, and Tal Schuster. 2022.
\newblock \href {https://doi.org/10.18653/v1/2022.emnlp-main.20} {Tomayto, tomahto. beyond token-level answer equivalence for question answering evaluation}.
\newblock In \emph{Proceedings of the 2022 Conference on Empirical Methods in Natural Language Processing}, pages 291--305, Abu Dhabi, United Arab Emirates. Association for Computational Linguistics.

\bibitem[{Chen et~al.(2021)Chen, Wang, and Wang}]{chen2021dataset}
Wenhu Chen, Xinyi Wang, and William~Yang Wang. 2021.
\newblock \href {http://arxiv.org/abs/2108.06314} {A dataset for answering time-sensitive questions}.

\bibitem[{Clark et~al.(2020)Clark, Luong, Le, and Manning}]{clark2020electra}
Kevin Clark, Minh-Thang Luong, Quoc~V. Le, and Christopher~D. Manning. 2020.
\newblock \href {https://openreview.net/forum?id=r1xMH1BtvB} {Electra: Pre-training text encoders as discriminators rather than generators}.
\newblock In \emph{International Conference on Learning Representations}.

\bibitem[{Crestani et~al.(1998)Crestani, Lalmas, Van~Rijsbergen, and Campbell}]{crestani1999bm25survey}
Fabio Crestani, Mounia Lalmas, Cornelis~J. Van~Rijsbergen, and Iain Campbell. 1998.
\newblock \href {https://doi.org/10.1145/299917.299920} {“is this document relevant?…probably”: A survey of probabilistic models in information retrieval}.
\newblock \emph{ACM Comput. Surv.}, 30(4):528–552.

\bibitem[{Dao(2023)}]{dao2023performance}
Xuan-Quy Dao. 2023.
\newblock \href {http://arxiv.org/abs/2307.02288} {Performance comparison of large language models on vnhsge english dataset: Openai chatgpt, microsoft bing chat, and google bard}.

\bibitem[{Datta et~al.(2022)Datta, Ganguly, Mitra, and Greene}]{querymetareview2}
Suchana Datta, Debasis Ganguly, Mandar Mitra, and Derek Greene. 2022.
\newblock \href {https://doi.org/10.1145/3545112} {A relative information gain-based query performance prediction framework with generated query variants}.
\newblock \emph{ACM Trans. Inf. Syst.}, 41(2).

\bibitem[{Devlin et~al.(2019)Devlin, Chang, Lee, and Toutanova}]{Devlin2019BERTPO}
Jacob Devlin, Ming-Wei Chang, Kenton Lee, and Kristina Toutanova. 2019.
\newblock \href {https://api.semanticscholar.org/CorpusID:52967399} {Bert: Pre-training of deep bidirectional transformers for language understanding}.
\newblock In \emph{North American Chapter of the Association for Computational Linguistics}.

\bibitem[{Dua et~al.(2019)Dua, Wang, Dasigi, Stanovsky, Singh, and Gardner}]{dua-etal-2019-drop}
Dheeru Dua, Yizhong Wang, Pradeep Dasigi, Gabriel Stanovsky, Sameer Singh, and Matt Gardner. 2019.
\newblock \href {https://doi.org/10.18653/v1/N19-1246} {{DROP}: A reading comprehension benchmark requiring discrete reasoning over paragraphs}.
\newblock In \emph{Proceedings of the 2019 Conference of the North {A}merican Chapter of the Association for Computational Linguistics: Human Language Technologies, Volume 1 (Long and Short Papers)}, pages 2368--2378, Minneapolis, Minnesota. Association for Computational Linguistics.

\bibitem[{Fan et~al.(2018)Fan, Guo, Lan, Xu, Zhai, and Cheng}]{10.1145/3209978.3209980}
Yixing Fan, Jiafeng Guo, Yanyan Lan, Jun Xu, Chengxiang Zhai, and Xueqi Cheng. 2018.
\newblock \href {https://doi.org/10.1145/3209978.3209980} {Modeling diverse relevance patterns in ad-hoc retrieval}.
\newblock In \emph{The 41st International ACM SIGIR Conference on Research \& Development in Information Retrieval}, SIGIR '18, page 375–384, New York, NY, USA. Association for Computing Machinery.

\bibitem[{Gabburo et~al.(2023)Gabburo, Garg, Kedziorski, and Moschitti}]{gabburo2023square}
Matteo Gabburo, Siddhant Garg, Rik~Koncel Kedziorski, and Alessandro Moschitti. 2023.
\newblock \href {http://arxiv.org/abs/2309.12250} {Square: Automatic question answering evaluation using multiple positive and negative references}.

\bibitem[{Gabburo et~al.(2022)Gabburo, Koncel-Kedziorski, Garg, Soldaini, and Moschitti}]{gabburo-etal-2022-knowledge}
Matteo Gabburo, Rik Koncel-Kedziorski, Siddhant Garg, Luca Soldaini, and Alessandro Moschitti. 2022.
\newblock \href {https://doi.org/10.18653/v1/2022.emnlp-main.645} {Knowledge transfer from answer ranking to answer generation}.
\newblock In \emph{Proceedings of the 2022 Conference on Empirical Methods in Natural Language Processing}, pages 9481--9495, Abu Dhabi, United Arab Emirates. Association for Computational Linguistics.

\bibitem[{Garg et~al.(2020)Garg, Vu, and Moschitti}]{Garg_Vu_Moschitti_2020}
Siddhant Garg, Thuy Vu, and Alessandro Moschitti. 2020.
\newblock \href {https://doi.org/10.1609/aaai.v34i05.6282} {Tanda: Transfer and adapt pre-trained transformer models for answer sentence selection}.
\newblock \emph{Proceedings of the AAAI Conference on Artificial Intelligence}, 34(05):7780--7788.

\bibitem[{Geva et~al.(2021)Geva, Khashabi, Segal, Khot, Roth, and Berant}]{geva2021strategyqa}
Mor Geva, Daniel Khashabi, Elad Segal, Tushar Khot, Dan Roth, and Jonathan Berant. 2021.
\newblock {Did Aristotle Use a Laptop? A Question Answering Benchmark with Implicit Reasoning Strategies}.
\newblock \emph{Transactions of the Association for Computational Linguistics (TACL)}.

\bibitem[{Gupta et~al.(2018)Gupta, Chinnakotla, and Shrivastava}]{gupta-etal-2018-retrieve}
Vishal Gupta, Manoj Chinnakotla, and Manish Shrivastava. 2018.
\newblock \href {https://doi.org/10.18653/v1/W18-5504} {Retrieve and re-rank: A simple and effective {IR} approach to simple question answering over knowledge graphs}.
\newblock In \emph{Proceedings of the First Workshop on Fact Extraction and {VER}ification ({FEVER})}, pages 22--27, Brussels, Belgium. Association for Computational Linguistics.

\bibitem[{He et~al.(2021)He, Gao, and Chen}]{he2021debertav3}
Pengcheng He, Jianfeng Gao, and Weizhu Chen. 2021.
\newblock \href {http://arxiv.org/abs/2111.09543} {Debertav3: Improving deberta using electra-style pre-training with gradient-disentangled embedding sharing}.

\bibitem[{Hsu et~al.(2021)Hsu, Lind, Soldaini, and Moschitti}]{hsu-etal-2021-answer}
Chao-Chun Hsu, Eric Lind, Luca Soldaini, and Alessandro Moschitti. 2021.
\newblock \href {https://doi.org/10.18653/v1/2021.findings-acl.374} {Answer generation for retrieval-based question answering systems}.
\newblock In \emph{Findings of the Association for Computational Linguistics: ACL-IJCNLP 2021}, pages 4276--4282, Online. Association for Computational Linguistics.

\bibitem[{Huang et~al.(2022)Huang, Geng, Long, and Jiang}]{huang-etal-2022-understand}
Hao Huang, Xiubo Geng, Guodong Long, and Daxin Jiang. 2022.
\newblock \href {https://doi.org/10.18653/v1/2022.naacl-main.28} {Understand before answer: Improve temporal reading comprehension via precise question understanding}.
\newblock In \emph{Proceedings of the 2022 Conference of the North American Chapter of the Association for Computational Linguistics: Human Language Technologies}, pages 375--384, Seattle, United States. Association for Computational Linguistics.

\bibitem[{Huang et~al.(2023)Huang, Yu, Ma, Zhong, Feng, Wang, Chen, Peng, Feng, Qin, and Liu}]{huang2023survey}
Lei Huang, Weijiang Yu, Weitao Ma, Weihong Zhong, Zhangyin Feng, Haotian Wang, Qianglong Chen, Weihua Peng, Xiaocheng Feng, Bing Qin, and Ting Liu. 2023.
\newblock \href {http://arxiv.org/abs/2311.05232} {A survey on hallucination in large language models: Principles, taxonomy, challenges, and open questions}.

\bibitem[{Jiang et~al.(2023)Jiang, Sablayrolles, Mensch, Bamford, Chaplot, de~Las~Casas, Bressand, Lengyel, Lample, Saulnier, Lavaud, Lachaux, Stock, Scao, Lavril, Wang, Lacroix, and Sayed}]{Jiang2023Mistral7}
Albert~Qiaochu Jiang, Alexandre Sablayrolles, Arthur Mensch, Chris Bamford, Devendra~Singh Chaplot, Diego de~Las~Casas, Florian Bressand, Gianna Lengyel, Guillaume Lample, Lucile Saulnier, L'elio~Renard Lavaud, Marie-Anne Lachaux, Pierre Stock, Teven~Le Scao, Thibaut Lavril, Thomas Wang, Timoth{\'e}e Lacroix, and William~El Sayed. 2023.
\newblock \href {https://api.semanticscholar.org/CorpusID:263830494} {Mistral 7b}.
\newblock \emph{ArXiv}, abs/2310.06825.

\bibitem[{Karpukhin et~al.(2020)Karpukhin, Oguz, Min, Lewis, Wu, Edunov, Chen, and Yih}]{karpukhin-etal-2020-dense}
Vladimir Karpukhin, Barlas Oguz, Sewon Min, Patrick Lewis, Ledell Wu, Sergey Edunov, Danqi Chen, and Wen-tau Yih. 2020.
\newblock \href {https://doi.org/10.18653/v1/2020.emnlp-main.550} {Dense passage retrieval for open-domain question answering}.
\newblock In \emph{Proceedings of the 2020 Conference on Empirical Methods in Natural Language Processing (EMNLP)}, pages 6769--6781, Online. Association for Computational Linguistics.

\bibitem[{Khattab and Zaharia(2020)}]{khattab2020colbert}
Omar Khattab and Matei Zaharia. 2020.
\newblock \href {https://doi.org/10.1145/3397271.3401075} {Colbert: Efficient and effective passage search via contextualized late interaction over bert}.
\newblock In \emph{Proceedings of the 43rd International ACM SIGIR Conference on Research and Development in Information Retrieval}, SIGIR '20, page 39–48, New York, NY, USA. Association for Computing Machinery.

\bibitem[{Kwiatkowski et~al.(2019)Kwiatkowski, Palomaki, Redfield, Collins, Parikh, Alberti, Epstein, Polosukhin, Kelcey, Devlin, Lee, Toutanova, Jones, Chang, Dai, Uszkoreit, Le, and Petrov}]{kwiatkowski2019nq}
Tom Kwiatkowski, Jennimaria Palomaki, Olivia Redfield, Michael Collins, Ankur Parikh, Chris Alberti, Danielle Epstein, Illia Polosukhin, Matthew Kelcey, Jacob Devlin, Kenton Lee, Kristina~N. Toutanova, Llion Jones, Ming-Wei Chang, Andrew Dai, Jakob Uszkoreit, Quoc Le, and Slav Petrov. 2019.
\newblock Natural questions: a benchmark for question answering research.
\newblock \emph{Transactions of the Association of Computational Linguistics}.

\bibitem[{Lewis et~al.(2020{\natexlab{a}})Lewis, Liu, Goyal, Ghazvininejad, Mohamed, Levy, Stoyanov, and Zettlemoyer}]{lewis-etal-2020-bart}
Mike Lewis, Yinhan Liu, Naman Goyal, Marjan Ghazvininejad, Abdelrahman Mohamed, Omer Levy, Veselin Stoyanov, and Luke Zettlemoyer. 2020{\natexlab{a}}.
\newblock \href {https://doi.org/10.18653/v1/2020.acl-main.703} {{BART}: Denoising sequence-to-sequence pre-training for natural language generation, translation, and comprehension}.
\newblock In \emph{Proceedings of the 58th Annual Meeting of the Association for Computational Linguistics}, pages 7871--7880, Online. Association for Computational Linguistics.

\bibitem[{Lewis et~al.(2020{\natexlab{b}})Lewis, Perez, Piktus, Petroni, Karpukhin, Goyal, K\"{u}ttler, Lewis, Yih, Rockt\"{a}schel, Riedel, and Kiela}]{lewis2020rag}
Patrick Lewis, Ethan Perez, Aleksandra Piktus, Fabio Petroni, Vladimir Karpukhin, Naman Goyal, Heinrich K\"{u}ttler, Mike Lewis, Wen-tau Yih, Tim Rockt\"{a}schel, Sebastian Riedel, and Douwe Kiela. 2020{\natexlab{b}}.
\newblock Retrieval-augmented generation for knowledge-intensive nlp tasks.
\newblock In \emph{Proceedings of the 34th International Conference on Neural Information Processing Systems}, NIPS'20, Red Hook, NY, USA. Curran Associates Inc.

\bibitem[{Lin and Ma(2021)}]{lin2021brief}
Jimmy Lin and Xueguang Ma. 2021.
\newblock \href {http://arxiv.org/abs/2106.14807} {A few brief notes on deepimpact, coil, and a conceptual framework for information retrieval techniques}.

\bibitem[{Liu et~al.(2019)Liu, Ott, Goyal, Du, Joshi, Chen, Levy, Lewis, Zettlemoyer, and Stoyanov}]{Liu2019RoBERTaAR}
Yinhan Liu, Myle Ott, Naman Goyal, Jingfei Du, Mandar Joshi, Danqi Chen, Omer Levy, Mike Lewis, Luke Zettlemoyer, and Veselin Stoyanov. 2019.
\newblock \href {https://api.semanticscholar.org/CorpusID:198953378} {Roberta: A robustly optimized bert pretraining approach}.
\newblock \emph{ArXiv}, abs/1907.11692.

\bibitem[{Luo et~al.(2023)Luo, Haihong, Tang, Peng, Guo, Zhang, Ma, Dong, Song, and Lin}]{Luo2023ChatKBQAAG}
Haoran Luo, E.~Haihong, Zichen Tang, Shiyao Peng, Yikai Guo, Wentai Zhang, Chenghao Ma, Guanting Dong, Meina Song, and Wei Lin. 2023.
\newblock \href {https://api.semanticscholar.org/CorpusID:264128221} {Chatkbqa: A generate-then-retrieve framework for knowledge base question answering with fine-tuned large language models}.
\newblock \emph{ArXiv}, abs/2310.08975.

\bibitem[{Mallia et~al.(2021)Mallia, Khattab, Tonellotto, and Suel}]{mallia2021learning}
Antonio Mallia, Omar Khattab, Nicola Tonellotto, and Torsten Suel. 2021.
\newblock \href {http://arxiv.org/abs/2104.12016} {Learning passage impacts for inverted indexes}.

\bibitem[{Mavi et~al.(2022)Mavi, Jangra, and Jatowt}]{Mavi2022ASO}
Vaibhav Mavi, Anubhav Jangra, and Adam Jatowt. 2022.
\newblock \href {https://api.semanticscholar.org/CorpusID:248266450} {A survey on multi-hop question answering and generation}.
\newblock \emph{ArXiv}, abs/2204.09140.

\bibitem[{Min et~al.(2019{\natexlab{a}})Min, Wallace, Singh, Gardner, Hajishirzi, and Zettlemoyer}]{min2019compositional}
Sewon Min, Eric Wallace, Sameer Singh, Matt Gardner, Hannaneh Hajishirzi, and Luke Zettlemoyer. 2019{\natexlab{a}}.
\newblock \href {http://arxiv.org/abs/1906.02900} {Compositional questions do not necessitate multi-hop reasoning}.

\bibitem[{Min et~al.(2019{\natexlab{b}})Min, Wallace, Singh, Gardner, Hajishirzi, and Zettlemoyer}]{min-etal-2019-compositional}
Sewon Min, Eric Wallace, Sameer Singh, Matt Gardner, Hannaneh Hajishirzi, and Luke Zettlemoyer. 2019{\natexlab{b}}.
\newblock \href {https://doi.org/10.18653/v1/P19-1416} {Compositional questions do not necessitate multi-hop reasoning}.
\newblock In \emph{Proceedings of the 57th Annual Meeting of the Association for Computational Linguistics}, pages 4249--4257, Florence, Italy. Association for Computational Linguistics.

\bibitem[{Nguyen et~al.(2016)Nguyen, Rosenberg, Song, Gao, Tiwary, Majumder, and Deng}]{nguyen2016ms}
Tri Nguyen, Mir Rosenberg, Xia Song, Jianfeng Gao, Saurabh Tiwary, Rangan Majumder, and Li~Deng. 2016.
\newblock \href {https://www.microsoft.com/en-us/research/publication/ms-marco-human-generated-machine-reading-comprehension-dataset/} {Ms marco: A human generated machine reading comprehension dataset}.

\bibitem[{Papineni et~al.(2001)Papineni, Roukos, Ward, and Zhu}]{papineni_bleu_2001}
Kishore Papineni, Salim Roukos, Todd Ward, and Wei-Jing Zhu. 2001.
\newblock \href {https://doi.org/10.3115/1073083.1073135} {{BLEU}: a method for automatic evaluation of machine translation}.
\newblock In \emph{Proceedings of the 40th Annual Meeting on Association for Computational Linguistics - {ACL} '02}, page 311. Association for Computational Linguistics.

\bibitem[{Perez et~al.(2020)Perez, Lewis, tau Yih, Cho, and Kiela}]{perez2020unsupervised}
Ethan Perez, Patrick Lewis, Wen tau Yih, Kyunghyun Cho, and Douwe Kiela. 2020.
\newblock \href {http://arxiv.org/abs/2002.09758} {Unsupervised question decomposition for question answering}.

\bibitem[{Petroni et~al.(2021)Petroni, Piktus, Fan, Lewis, Yazdani, De~Cao, Thorne, Jernite, Karpukhin, Maillard, Plachouras, Rockt{\"a}schel, and Riedel}]{petroni-etal-2021-kilt}
Fabio Petroni, Aleksandra Piktus, Angela Fan, Patrick Lewis, Majid Yazdani, Nicola De~Cao, James Thorne, Yacine Jernite, Vladimir Karpukhin, Jean Maillard, Vassilis Plachouras, Tim Rockt{\"a}schel, and Sebastian Riedel. 2021.
\newblock \href {https://doi.org/10.18653/v1/2021.naacl-main.200} {{KILT}: a benchmark for knowledge intensive language tasks}.
\newblock In \emph{Proceedings of the 2021 Conference of the North American Chapter of the Association for Computational Linguistics: Human Language Technologies}, pages 2523--2544, Online. Association for Computational Linguistics.

\bibitem[{Radford et~al.(2018)Radford, Narasimhan, Salimans, and Sutskever}]{radford2018improving}
Alec Radford, Karthik Narasimhan, Tim Salimans, and Ilya Sutskever. 2018.
\newblock Improving language understanding by generative pre-training.

\bibitem[{Raffel et~al.(2020)Raffel, Shazeer, Roberts, Lee, Narang, Matena, Zhou, Li, and Liu}]{Raffel2020t5}
Colin Raffel, Noam Shazeer, Adam Roberts, Katherine Lee, Sharan Narang, Michael Matena, Yanqi Zhou, Wei Li, and Peter~J. Liu. 2020.
\newblock \href {http://jmlr.org/papers/v21/20-074.html} {Exploring the limits of transfer learning with a unified text-to-text transformer}.
\newblock \emph{Journal of Machine Learning Research}, 21(140):1--67.

\bibitem[{Rogers et~al.(2023)Rogers, Gardner, and Augenstein}]{10.1145/3560260}
Anna Rogers, Matt Gardner, and Isabelle Augenstein. 2023.
\newblock \href {https://doi.org/10.1145/3560260} {Qa dataset explosion: A taxonomy of nlp resources for question answering and reading comprehension}.
\newblock \emph{ACM Comput. Surv.}, 55(10).

\bibitem[{Saxena et~al.(2021)Saxena, Chakrabarti, and Talukdar}]{saxena-etal-2021-question}
Apoorv Saxena, Soumen Chakrabarti, and Partha Talukdar. 2021.
\newblock \href {https://doi.org/10.18653/v1/2021.acl-long.520} {Question answering over temporal knowledge graphs}.
\newblock In \emph{Proceedings of the 59th Annual Meeting of the Association for Computational Linguistics and the 11th International Joint Conference on Natural Language Processing (Volume 1: Long Papers)}, pages 6663--6676, Online. Association for Computational Linguistics.

\bibitem[{Shang et~al.(2022)Shang, Wang, Qi, and Huang}]{shang-etal-2022-improving}
Chao Shang, Guangtao Wang, Peng Qi, and Jing Huang. 2022.
\newblock \href {https://doi.org/10.18653/v1/2022.acl-long.552} {Improving time sensitivity for question answering over temporal knowledge graphs}.
\newblock In \emph{Proceedings of the 60th Annual Meeting of the Association for Computational Linguistics (Volume 1: Long Papers)}, pages 8017--8026, Dublin, Ireland. Association for Computational Linguistics.

\bibitem[{Sharma et~al.(2023)Sharma, Saxena, Gupta, Kazemi, Talukdar, and Chakrabarti}]{sharma-etal-2023-twirgcn}
Aditya Sharma, Apoorv Saxena, Chitrank Gupta, Mehran Kazemi, Partha Talukdar, and Soumen Chakrabarti. 2023.
\newblock \href {https://aclanthology.org/2023.eacl-main.150} {{T}wi{RGCN}: Temporally weighted graph convolution for question answering over temporal knowledge graphs}.
\newblock In \emph{Proceedings of the 17th Conference of the European Chapter of the Association for Computational Linguistics}, pages 2049--2060, Dubrovnik, Croatia. Association for Computational Linguistics.

\bibitem[{Su et~al.(2022)Su, Lee, Hsu, and Su}]{su-etal-2022-roberta}
Ming-Hsiang Su, Chin-Wei Lee, Chi-Lun Hsu, and Ruei-Cyuan Su. 2022.
\newblock \href {https://aclanthology.org/2022.rocling-1.8} {{R}o{BERT}a-based traditional {C}hinese medicine named entity recognition model}.
\newblock In \emph{Proceedings of the 34th Conference on Computational Linguistics and Speech Processing (ROCLING 2022)}, pages 61--66, Taipei, Taiwan. The Association for Computational Linguistics and Chinese Language Processing (ACLCLP).

\bibitem[{Sun et~al.(2022)Sun, He, Qiu, and Huang}]{sun-etal-2022-bertscore}
Tianxiang Sun, Junliang He, Xipeng Qiu, and Xuanjing Huang. 2022.
\newblock \href {https://doi.org/10.18653/v1/2022.emnlp-main.245} {{BERTS}core is unfair: On social bias in language model-based metrics for text generation}.
\newblock In \emph{Proceedings of the 2022 Conference on Empirical Methods in Natural Language Processing}, pages 3726--3739, Abu Dhabi, United Arab Emirates. Association for Computational Linguistics.

\bibitem[{Talmor and Berant(2018)}]{talmor18compwebq}
A.~Talmor and J.~Berant. 2018.
\newblock The web as a knowledge-base for answering complex questions.
\newblock In \emph{North American Association for Computational Linguistics (NAACL)}.

\bibitem[{Trivedi et~al.(2022)Trivedi, Balasubramanian, Khot, and Sabharwal}]{trivedi2022musique}
Harsh Trivedi, Niranjan Balasubramanian, Tushar Khot, and Ashish Sabharwal. 2022.
\newblock \href {https://doi.org/10.1162/tacl_a_00475} {{M}u{S}i{Q}ue: Multihop questions via single-hop question composition}.
\newblock \emph{Transactions of the Association for Computational Linguistics}, 10:539--554.

\bibitem[{Vaswani et~al.(2017)Vaswani, Shazeer, Parmar, Uszkoreit, Jones, Gomez, Kaiser, and Polosukhin}]{vasawani2017}
Ashish Vaswani, Noam Shazeer, Niki Parmar, Jakob Uszkoreit, Llion Jones, Aidan~N. Gomez, \L{}ukasz Kaiser, and Illia Polosukhin. 2017.
\newblock Attention is all you need.
\newblock In \emph{Proceedings of the 31st International Conference on Neural Information Processing Systems}, NIPS'17, page 6000–6010, Red Hook, NY, USA. Curran Associates Inc.

\bibitem[{Vu and Moschitti(2021)}]{vu-moschitti-2021-ava}
Thuy Vu and Alessandro Moschitti. 2021.
\newblock \href {https://doi.org/10.18653/v1/2021.naacl-main.412} {{AVA}: an automatic e{V}aluation approach for question answering systems}.
\newblock In \emph{Proceedings of the 2021 Conference of the North American Chapter of the Association for Computational Linguistics: Human Language Technologies}, pages 5223--5233, Online. Association for Computational Linguistics.

\bibitem[{Vu et~al.(2023)Vu, Iyyer, Wang, Constant, Wei, Wei, Tar, Sung, Zhou, Le, and Luong}]{vu2023freshllms}
Tu~Vu, Mohit Iyyer, Xuezhi Wang, Noah Constant, Jerry Wei, Jason Wei, Chris Tar, Yun-Hsuan Sung, Denny Zhou, Quoc Le, and Thang Luong. 2023.
\newblock \href {http://arxiv.org/abs/2310.03214} {Freshllms: Refreshing large language models with search engine augmentation}.

\bibitem[{Wang et~al.(2020)Wang, Fan, Guo, Yang, Zhang, Lan, Cheng, Jiang, and Wang}]{wang2020match}
Zizhen Wang, Yixing Fan, Jiafeng Guo, Liu Yang, Ruqing Zhang, Yanyan Lan, Xueqi Cheng, Hui Jiang, and Xiaozhao Wang. 2020.
\newblock \href {https://doi.org/10.1145/3397271.3401143} {Match²: A matching over matching model for similar question identification}.
\newblock In \emph{Proceedings of the 43rd International ACM SIGIR Conference on Research and Development in Information Retrieval}, SIGIR '20, page 559–568, New York, NY, USA. Association for Computing Machinery.

\bibitem[{Wei et~al.(2022)Wei, Wang, Schuurmans, Bosma, Ichter, Xia, Chi, Le, and Zhou}]{wei2022chainofthought}
Jason Wei, Xuezhi Wang, Dale Schuurmans, Maarten Bosma, Brian Ichter, Fei Xia, Ed~Chi, Quoc Le, and Denny Zhou. 2022.
\newblock \href {http://arxiv.org/abs/2201.11903} {Chain-of-thought prompting elicits reasoning in large language models}.
\newblock Cite arxiv:2201.11903.

\bibitem[{Yan et~al.(2023)Yan, Wang, Zhao, Huang, Chen, and Wang}]{yan-etal-2023-bleurt}
Yiming Yan, Tao Wang, Chengqi Zhao, Shujian Huang, Jiajun Chen, and Mingxuan Wang. 2023.
\newblock \href {https://doi.org/10.18653/v1/2023.acl-long.297} {{BLEURT} has universal translations: An analysis of automatic metrics by minimum risk training}.
\newblock In \emph{Proceedings of the 61st Annual Meeting of the Association for Computational Linguistics (Volume 1: Long Papers)}, pages 5428--5443, Toronto, Canada. Association for Computational Linguistics.

\bibitem[{Yang et~al.(2018{\natexlab{a}})Yang, Qi, Zhang, Bengio, Cohen, Salakhutdinov, and Manning}]{yang-etal-2018-hotpotqa}
Zhilin Yang, Peng Qi, Saizheng Zhang, Yoshua Bengio, William Cohen, Ruslan Salakhutdinov, and Christopher~D. Manning. 2018{\natexlab{a}}.
\newblock \href {https://doi.org/10.18653/v1/D18-1259} {{H}otpot{QA}: A dataset for diverse, explainable multi-hop question answering}.
\newblock In \emph{Proceedings of the 2018 Conference on Empirical Methods in Natural Language Processing}, pages 2369--2380, Brussels, Belgium. Association for Computational Linguistics.

\bibitem[{Yang et~al.(2018{\natexlab{b}})Yang, Qi, Zhang, Bengio, Cohen, Salakhutdinov, and Manning}]{yang2018hotpotqa}
Zhilin Yang, Peng Qi, Saizheng Zhang, Yoshua Bengio, William~W. Cohen, Ruslan Salakhutdinov, and Christopher~D. Manning. 2018{\natexlab{b}}.
\newblock {HotpotQA}: A dataset for diverse, explainable multi-hop question answering.
\newblock In \emph{Conference on Empirical Methods in Natural Language Processing ({EMNLP})}.

\bibitem[{Yoran et~al.(2023)Yoran, Wolfson, Bogin, Katz, Deutch, and Berant}]{yoran2023answering}
Ori Yoran, Tomer Wolfson, Ben Bogin, Uri Katz, Daniel Deutch, and Jonathan Berant. 2023.
\newblock \href {http://arxiv.org/abs/2304.13007} {Answering questions by meta-reasoning over multiple chains of thought}.

\bibitem[{Zhou and Croft(2007)}]{querymetareview1}
Yun Zhou and W.~Bruce Croft. 2007.
\newblock \href {https://doi.org/10.1145/1277741.1277835} {Query performance prediction in web search environments}.
\newblock In \emph{Proceedings of the 30th Annual International ACM SIGIR Conference on Research and Development in Information Retrieval}, SIGIR '07, page 543–550, New York, NY, USA. Association for Computing Machinery.

\end{thebibliography}
